\documentclass[letterpaper]{article} 
\usepackage{aaai2026}  
\usepackage{times}  
\usepackage{helvet}  
\usepackage{courier}  
\usepackage[hyphens]{url}  
\usepackage{graphicx} 
\usepackage{natbib}  
\usepackage{caption} 
\usepackage{booktabs}
\usepackage{algorithm}      
\usepackage{algpseudocode}  
\usepackage{amsmath}
\usepackage{subcaption}
\usepackage{pifont}
\usepackage{amsfonts}
\usepackage{multirow} 
 
\usepackage{newfloat}
\usepackage{listings}

\usepackage[dvipsnames,x11names]{xcolor}
\newcommand{\cmark}{\ding{51}} 
\newcommand{\xmark}{\ding{55}} 
\definecolor{AliGreen}{HTML}{2CA02C}   
\definecolor{AliBrown}{HTML}{000000}   

\newcommand{\ali}[1]{\textcolor{AliBrown}{#1}}   

\newcommand{\draft}[2]{\textcolor{#1}{#2}}
\newcommand{\salmaan}[1]  {\draft{black}{#1}}

\newcommand{\salman}[1]  {\draft{black}{#1}}
\newcommand{\tahir}[1] {\draft{black}{#1}}

\DeclareCaptionStyle{ruled}{labelfont=normalfont,labelsep=colon,strut=off} 
\floatstyle{ruled}
\newfloat{listing}{tb}{lst}{}
\floatname{listing}{Listing}
\title{I-INR: Iterative Implicit Neural Representations}

\author{
Ali Haider\textsuperscript{\rm 1},
Muhammad Salman Ali\textsuperscript{\rm 1},
Maryam Qamar\textsuperscript{\rm 1},
Tahir Khalil\textsuperscript{\rm 1},
Soo Ye Kim\textsuperscript{\rm 2},
Jihyong Oh\textsuperscript{\rm 3}\thanks{Corresponding Authors}\!,
Enzo Tartaglione\textsuperscript{\rm 4},
Sung\mbox{-}Ho Bae\textsuperscript{\rm 1}\footnotemark[1]
}

\affiliations{
\textsuperscript{\rm 1}Kyung Hee University, Republic of Korea\\
\textsuperscript{\rm 2}Adobe Research\\
\textsuperscript{\rm 3}Chung-Ang University, Republic of Korea\\
\textsuperscript{\rm 4}LTCI, T\'el\'ecom Paris, Institut Polytechnique de Paris, France\\
\{alihaider, salmanali, maryamqamar, tahirkhalil26\}@khu.ac.kr, sooyek@adobe.com,\\
jihyongoh@cau.ac.kr, enzo.tartaglione@telecom-paris.fr, shbae@khu.ac.kr
}

\begin{document}

\maketitle

\begin{abstract}
\ali{Implicit Neural Representations (INRs) have revolutionized signal processing and computer vision by modeling signals as continuous, differentiable functions parameterized by neural networks. However, INRs are prone to the spectral bias problem, limiting their ability to retain high-frequency information, and often struggle with noise robustness. Motivated by recent trends in iterative refinement processes, we propose Iterative Implicit Neural Representations (I-INRs). This novel plug-and-play framework iteratively refines signal reconstructions to restore high-frequency details, improve noise robustness, and enhance generalization, ultimately delivering superior reconstruction quality. I-INRs integrate seamlessly into existing INR architectures with only a 0.5–2\% increase in parameters. During reconstruction, the iterative refinement adds just 0.8–1.6\% additional FLOPs over the baseline while delivering a substantial performance boost of up to +2.0 PSNR. Extensive experiments demonstrate that I-INRs consistently outperform WIRE, SIREN, and Gauss across various computer vision tasks, including image fitting, image denoising, and object occupancy prediction.
The code is available at \url{https://github.com/optimizer077/I-INR}.}
\end{abstract}    
\begin{figure}[t]
    \centering
    \includegraphics[width=\linewidth]{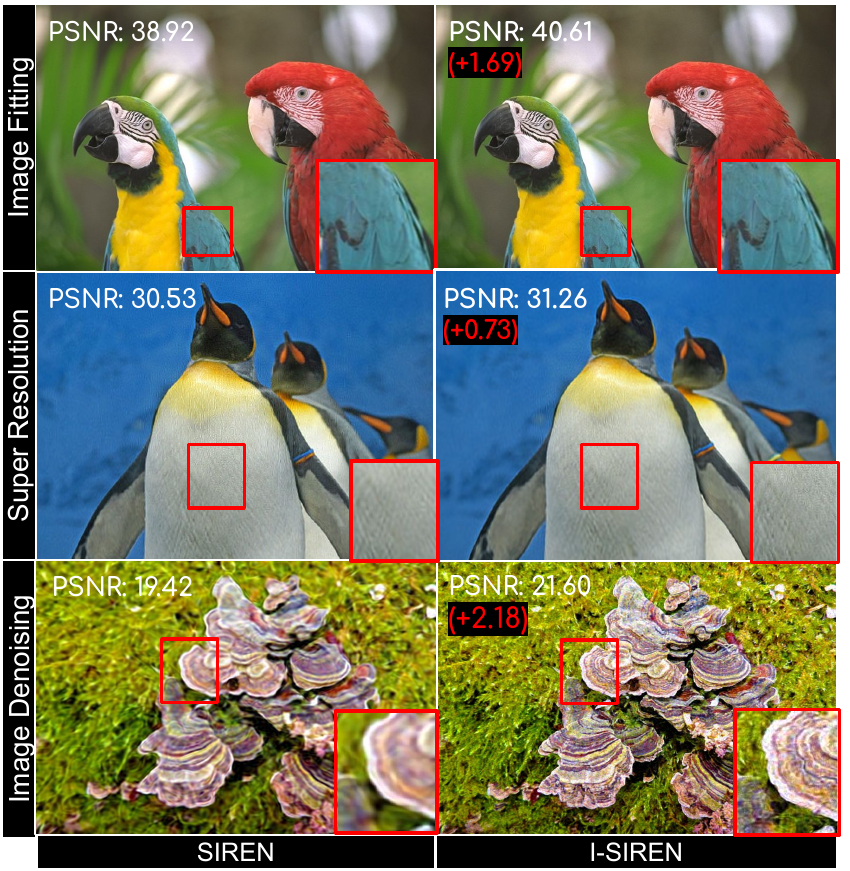}
    \caption{
    Effectiveness of the proposed methods across multiple tasks, including image fitting, super resolution (2$\times$ scale), and denoising, compared to the baseline representative INR method (SIREN). Our novel Iterative-INR method (I-SIREN) consistently improves detail preservation, fidelity, and high-frequency reconstruction across all tasks, outperforming the baseline.}
    \label{fig:teaser}
\end{figure}

\section{Introduction}
Implicit Neural Representations (INRs) have emerged as a transformative approach in signal representation, shifting from traditional discrete grid-based methods to continuous coordinate-based models \cite{sitzmann2020implicit}. By leveraging neural networks, typically multi-layer perceptrons (MLPs), INRs map spatial or temporal coordinates directly to signal attributes such as pixel intensity, colour, or 3D occupancy. This continuous representation inherently offers resolution independence, compact encoding, and seamless interpolation, making INRs highly versatile across a wide range of domains, including computer vision \cite{chen2021learning, saragadam2022miner, sitzmann2020implicit, mildenhall2021nerf, jiang2023coordinate}, and beyond \cite{sun2021coil, shen2021non, zhong2019reconstructing, reed2021implicit}. Initially introduced for tasks like 3D shape reconstruction \cite{mildenhall2021nerf, peng2020convolutional, srinivasan2021nerv} and novel view synthesis \cite{Irshad_2023_ICCV}, INRs have since been applied to diverse applications including image fitting~\cite{ liu2023finer}, super-resolution~\cite{delbracioinversion, saharia2022image}, and solving inverse problems~\cite{cha2024descanning,wang2019spatial,chen2021learning}.

\salmaan{Despite advancements, INRs continue to face substantial challenges in capturing high-frequency details, maintaining robustness against noise, and handling incomplete data~\cite{saragadam2023wire}. 
Typically optimized with L1 or L2 loss~\cite{sitzmann2020implicit}, INRs inherently exhibit a spectral bias that favors low-frequency components, often at the expense of fine-grained details~\cite{rahaman2019spectral}.}

To mitigate this, recent approaches have introduced positional encoding schemes~\cite{tancik2020fourier,fathony2020multiplicative} and alternative activation functions~\cite{sitzmann2020implicit,saragadam2023wire,liu2023finer, gao2024h, thennakoon2025bandrc} to better capture high-frequency content. Positional encodings inject high-frequency signals through orthogonal Fourier bases~\cite{tancik2020fourier}, while periodic activations such as sine functions enable MLPs to model high-frequency structures more effectively compared to traditional ReLU-based networks~\cite{sitzmann2020implicit}. Nevertheless, these methods often assume clean and fully observed inputs, limiting their effectiveness in practical scenarios characterized by noise and occlusion~\cite{saragadam2023wire}. More recent efforts, such as the integration of complex Gabor wavelet activations, have aimed to improve robustness under such conditions~\cite{saragadam2023wire}, but challenges persist when scaling to real-world, noisy data. Consequently, there remains significant potential to enhance the fidelity, robustness, and generalization of INRs, particularly in preserving high-frequency signals under adverse conditions.


\ali{To address these limitations, and drawing inspiration from iterative models~\cite{delbracioinversion, rissanen2022generative, ho2020denoising}, we introduce \textit{Iterative Implicit Neural Representations (I-INRs)} a plug-and-play framework that integrates seamlessly with existing implicit architectures. Unlike traditional single-shot INRs, I-INRs reconstruct signals progressively over multiple iterations, refining details, enhancing reconstruction quality, and improving noise robustness, as illustrated in Figure~\ref{fig:teaser}.} 

\begin{figure}[t]
    \centering
    \subfloat[]{
        \includegraphics[width=\columnwidth]{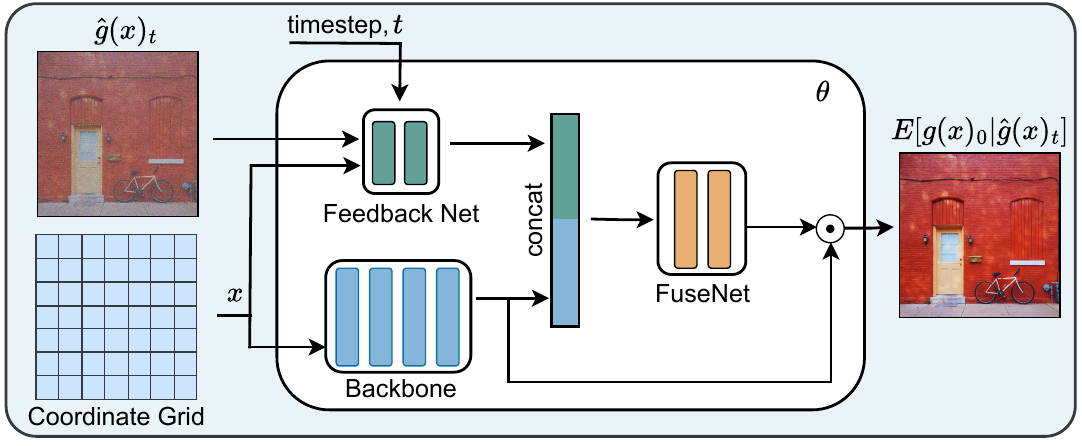}
        \label{fig:architecture}
    } \\
    \vfill
    \subfloat[]{
        \includegraphics[width=\columnwidth]{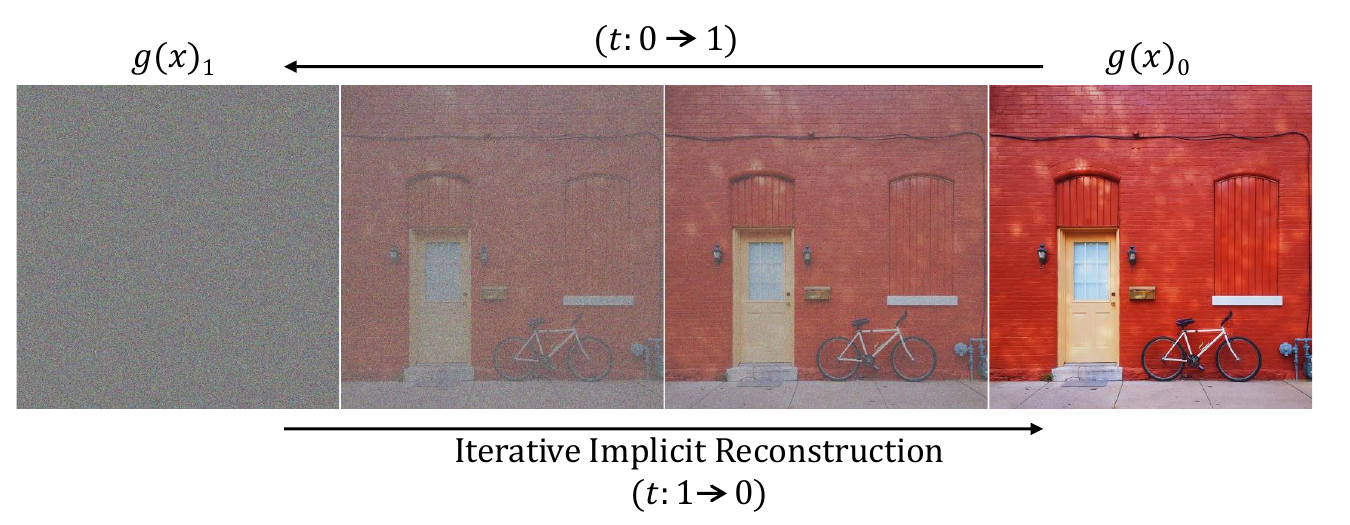}
        \label{fig:iterative_process}
    }
    \caption{
    (a) The proposed architecture of the I-INR model. The framework consists of a  Backbone, which can be any baseline INR architecture. Additionally, a Feedback Net incorporates feedback to refine representations, while FuseNet integrates features. The final output is obtained by combining the outputs of the Backbone and the FuseNet, enhancing expressivity and reconstruction quality. (b) Iterative reconstruction process of the proposed I-INR framework.}
    \label{fig:arch_iterative_recon}
\end{figure}

The proposed framework features a novel architecture that incorporates existing implicit networks as a BackboneNet without major modifications, while introducing two lightweight modules, i.e FeedbackNet and FuseNet as shown in Figure~\ref{fig:architecture}. These blocks add only 0.5–2\% to the total parameter count of the baseline model and support the iterative reconstruction process. Owing to this design, the iterative I-INR process incurs only 0.8–2\% additional FLOPs over the baseline INR for two-step reconstruction. Our experiments demonstrate that I-INRs deliver superior reconstructions with enhanced detail, better generalization, and minimal computational overhead, consistently outperforming single-shot methods such as WIRE~\cite{saragadam2023wire}, SIREN~\cite{sitzmann2020implicit}, and Gauss~\cite{ramasinghe2022beyond} across tasks including image fitting, super-resolution, denoising, and 3D occupancy reconstruction.

Our main contributions are as follows:


\begin{itemize}
    \item We present \textbf{I}terative \textbf{I}mplicit \textbf{N}eural \textbf{R}epresentations, called \textbf{I-INRs}, a novel framework for INRs that reconstructs signals \textit{iteratively}, effectively capturing high-frequency details, enhancing robustness to noise, and improving generalization.
    \item I-INRs are designed as a plug-and-play framework, compatible with existing implicit architectures, enabling seamless integration and adoption in various applications.
    \item We validate the effectiveness of I-INRs through extensive experiments across multiple tasks, demonstrating superior performance over traditional single-shot INRs in terms of detail reconstruction and robustness.
\end{itemize}

\section{Related Work}
\label{Related_Work}
\noindent
\paragraph{Spatial Encoding Techniques.}
INRs suffer from spectral bias~\cite{rahaman2019spectral,tancik2020fourier}, struggling to capture high-fidelity details of complex signals. This limitation affects applications requiring fine-grained detail, such as a single image super resolution (SR), 3D reconstruction, and scene modeling. To address this challenge, various approaches have been proposed, including spatial encoding with frequency-based representations, polynomial decomposition~\cite{raghavan2023neural,singh2023polynomial}, and high-pass filtering~\cite{fathony2020multiplicative}, all of which emphasize high-frequency components. DINER~\cite{xie2023diner} applies a hash map to unevenly map input coordinates to feature vectors, optimizing spatial frequency distribution for faster, more accurate reconstructions.


\noindent
\paragraph{Role of Activations.}
Beyond encoding, activation functions critically influence spectral bias. Standard choices like ReLU cannot capture high-frequency components due to limited representational capacity~\cite{sitzmann2020implicit}. To address this, alternatives such as sinusoidal~\cite{sitzmann2020implicit} and Gaussian~\cite{ramasinghe2022beyond} activations have been proposed. Activations tuned for high accuracy can suffer from reduced noise robustness; WIRE ~\cite{saragadam2023wire} leverages Gabor wavelets to provide enhanced robustness to noise. However, there remains room for improvement in achieving both high-frequency fitting and noise robustness ~\cite{saragadam2023wire}.

\noindent
\paragraph{Iterative Methods.} Iterative models have been very successful in image restoration and translation~\cite{saharia2022image, rissanen2022generative, delbracioinversion, hui2024microdiffusion, chu2025highly,chen2024image}, video generation~\cite{ji2025layerflow}, language modeling~\cite{nie2025large}, 3D generation~\cite{qian2023magic123, liu2023zero}, and more. In contrast with single-shot models, iterative models such as diffusion models~\cite{ho2020denoising,rissanen2022generative,delbracioinversion} produce high-quality samples by reversing degradation processes over multiple steps. Despite their widespread adoption and effectiveness, iterative models remain underexplored in the context of implicit neural representations.

Inspired by the strong detail reconstruction capabilities of iterative models, we propose a novel INR framework that reconstructs signals implicitly over multiple steps. Our approach provides improved details and enhanced noise robustness, offering a flexible, plug-and-play solution that can be integrated into any existing INR framework. However, for the scope of this paper, we specifically focus on WIRE~\cite{saragadam2023wire}, Gauss~\cite{ramasinghe2022beyond}, and SIREN~\cite{sitzmann2020implicit}.

\section{Preliminary}

\subsection{Standard INR}
Let $g : X \subseteq \mathbb{R}^p \rightarrow Y \subseteq \mathbb{R}^q$ be a target function mapping $p$-dimensional inputs (such as pixel positions in a 2D image or volumetric domains in 3D scenes) to $q$-dimensional outputs (such as color or density values), such that $y = g(x)$ for $x \in X$. An INR~\cite{sitzmann2020implicit} is a learnable neural function $f_\theta : X \rightarrow Y$, parameterized by $\theta$, aiming to approximate $\mathcal{I}(x)$ across $X$. The objective is to find $\theta$ such that:
\begin{equation}
f_{\theta}(x) \approx \mathcal{I}(x),
\end{equation}
where $\mathcal{I}(x)$ represents a true image function evaluated at coordinates $x$. A typical INR is trained by reducing the signal error in an $L_p$ metric:
\begin{equation}
\min_{\theta}\mathbb{E}_{x}\left[\|f_{\theta}(x)-\mathcal{I}(x)\|_p\right]\approx \min_{\theta}\sum_i \|f_{\theta}(x^i)-\mathcal{I}(x^i)\|_p,
\end{equation}
where $p = 2$ for standard INR training, corresponding to the mean squared error.

\subsection{Inversion by Direct Iteration (InDI)}

Inversion by Direct Iteration (InDI)~\cite{delbracioinversion} addresses ill-posed inverse problems, such as image restoration, by iteratively refining an estimate of the unknown signal \(b\) from an observed measurement \(a\). Instead of solving the full inverse problem in one step, InDI interpolates between \(a\) and \(b\) via \(b_t = t a + (1-t)b\), with \(t \in [0, 1]\), and uses a neural network \(F_\theta\) to approximate the conditional expectation \( \mathbb{E}[b \mid b_t] \). The network is trained to minimize:
\[
\min_\theta\, \mathbb{E}_{a, b, t, n}\left[\, \| F_\theta(b_t + \epsilon_t n,\, t) - b \|_1 \,\right],
\]
where \(\epsilon_t n\) is Gaussian noise. At inference, the estimate is updated recursively using the learned conditional mean, gradually solving a sequence of easier inverse problems from \(a\) to \(b\).
\section{Proposed Method}
In this section, we present a detailed explanation of our proposed methods, including a comprehensive description of the iterative plug-and-play framework.
\label{IINR}

\subsection{Iterative Implicit Neural Representations: I-INR}
\label{i-inr-formulation}




\begin{figure*}[t]
    \centering
    \includegraphics[width=\linewidth]{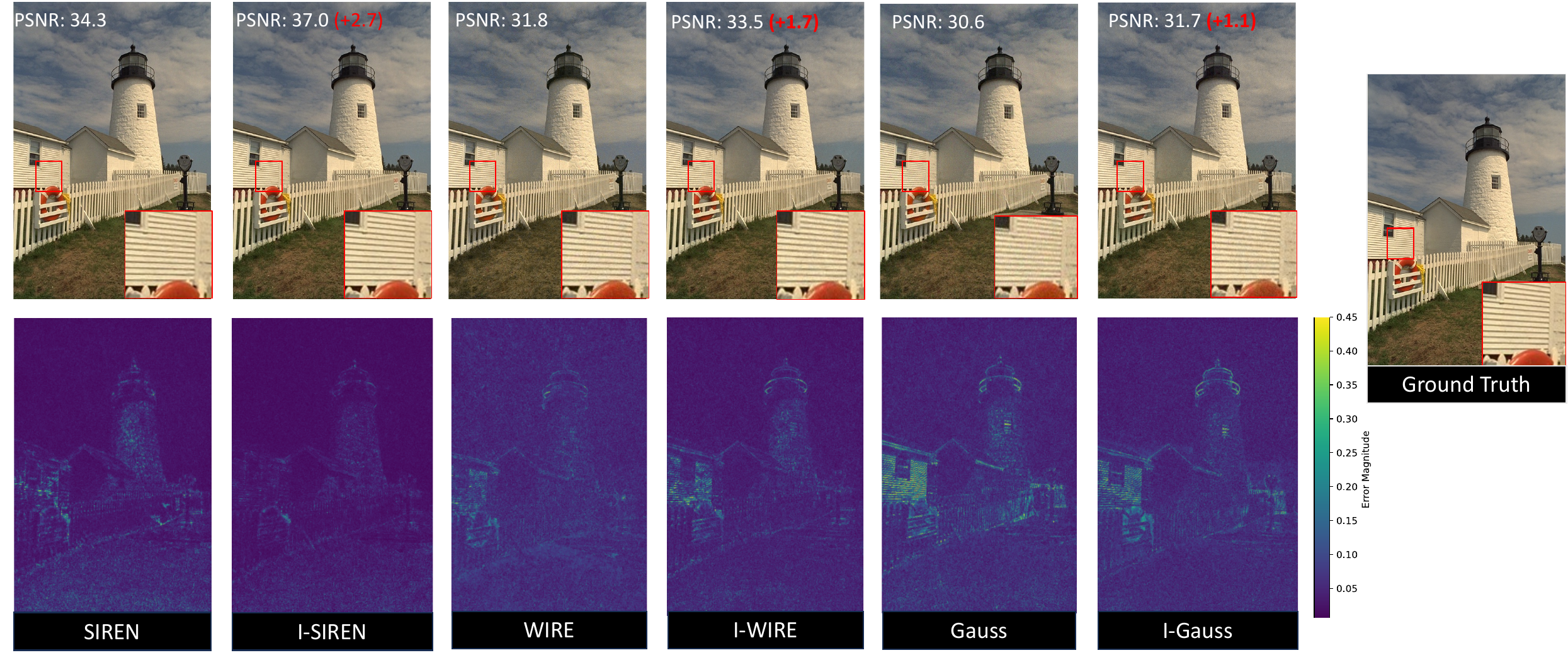}
    \caption{Image fitting results for different non-linearities and their iterative extensions (prefix "I"). Each column presents the reconstructed image (top) and residual map (bottom), highlighting reconstruction quality. PSNR values are reported, with iterative improvements in red. The rightmost column shows the ground truth.}
    \label{fig:Image_Fitting}
\end{figure*}






I-INR reconstructs the signal over multiple steps, progressively improving the quality at each iteration. As shown in Figure~\ref{fig:iterative_process}, the process starts from an initial state \( g(x)_1 = \mathcal{Z} \) at \( t = 1 \) and gradually approaches the final reconstruction \( g(x)_0 = \mathcal{I}(x) \) at \( t = 0 \), using steps of size \( \delta \). This is achieved by inverting the forward process:
\begin{equation}
g(x)_t = \mathcal{I}(x)(1 - t) + \mathcal{Z} t, \qquad t \in [0, 1],
\label{IINR}
\end{equation}
where \( \mathcal{Z} \) is sampled from a known distribution matching the shape of \( \mathcal{I}(x) \).

At each step, I-INR estimates the next state by computing the conditional expectation \( \mathbb{E}[g(x)_0 \mid g(x)_t] \), which represents the best possible prediction of the original signal given the current state. This estimate is then used to update the reconstruction as follows:
\begin{equation}
\begin{aligned}
\hat{g}(x)_{t-\delta}
    &= \mathbb{E}\left[ g(x)_{t-\delta} \mid \hat{g}(x)_t \right] \\
    &= \frac{\delta}{t} \, \mathbb{E}\left[ g(x)_0 \mid \hat{g}(x)_t \right]
    + \left( 1 - \frac{\delta}{t} \right) \hat{g}(x)_t,
\label{Recons}
\end{aligned}
\end{equation}
by repeatedly applying this update, I-INR smoothly transitions from the initial state to the reconstructed signal, leveraging the conditional expectation at each step to improve accuracy.

\paragraph{I--INR Training.}
To realize the update rule in Eq.~(\ref{Recons}), we train an implicit neural network \(f_{\theta}\) (see Figure.~\ref{fig:architecture}) that maps the intermediate state \(\tilde{g}(x)_t\), the spatial coordinates \(x\), and the time index \(t\) directly to the clean target \(\mathcal{I}(x)\):

\begin{align}
\min_{\theta}\;
\mathbb{E}_{x,\,t,\,n}
\left\|
    f_{\theta}\!\bigl(\tilde g(x)_t,\,x,\,t\bigr)
    - \mathcal{I}(x)
\right\|_{2}^{2},
\label{eq:training_obj}
\\[4pt] 
\tilde g(x)_t
= (1 - t)\,\mathcal{I}(x)
+ t\,\mathcal{Z}
+ \varepsilon t n,
\label{eq:noise_schedule}
\end{align}

\noindent
where \(\mathcal{Z}\sim p(\mathcal{Z})\) and \(n\sim\mathcal{N}(0,1)\).
The small perturbation \(\varepsilon t n\) \cite{delbracioinversion}
satisfies the regularity requirement, guaranteeing stable reconstruction
during inference. The full training procedure appears in Algorithm~\ref{algo1}.

\paragraph{I--INR Reconstruction.}
Reconstruction begins from the latent state $\mathcal{Z}$ and proceeds for
$1/\delta$ iterations of Eq.\,~\ref{Recons}, iteratively refining the
estimate towards the final signal (Algorithm~\ref{algo2}).

\begin{algorithm}[ht]
  \caption{I-INRs Model Training of $f_\theta$}
  \begin{algorithmic}[1]
    \Require $\mathcal{I}(x)$, $\varepsilon$, $\eta$
    \State $\mathcal{Z} \sim p(\mathcal{Z})$
    \Repeat
      \State $x \sim p(x)$ 
      \State $t \sim \mathcal{U}(0,1),\; n \sim \mathcal{N}(0,I)$
      \State $\tilde g(x)_t \gets (1-t)\,\mathcal{I}(x) + t\,\mathcal{Z} + \varepsilon\,t\,n$
      \State $\displaystyle
        \theta \gets \theta - \eta\,
        \nabla_{\theta}
        \bigl\| f_\theta(\tilde g(x)_t,\,x,\,t) - \mathcal{I}(x) \bigr\|_2^{2}$
    \Until{\textbf{converged}}
  \end{algorithmic}
  \label{algo1}
\end{algorithm}

\begin{algorithm}[ht]
  \caption{I-INRs Model Reconstruction}
  \begin{algorithmic}[1]
    \Require $\mathcal{Z}$, $f_\theta$, $\delta$, $x$
    \State $\hat g(x)_1 \gets \mathcal{Z}$
    \For{$t \gets 1$ \textbf{to} $0$ \textbf{step} $-\delta$}
      \State $\hat g(x)_{t-\delta} \gets
             \dfrac{\delta}{t}\, f_\theta\bigl(\hat g(x)_t, x, t\bigr) +
             \left(1 - \dfrac{\delta}{t}\right) \hat g(x)_t$
    \EndFor
    \State \Return $\hat g(x)_0$
  \end{algorithmic}
  \label{algo2}
\end{algorithm}

\begin{figure*}[t]
    \centering
    \includegraphics[width=\linewidth]{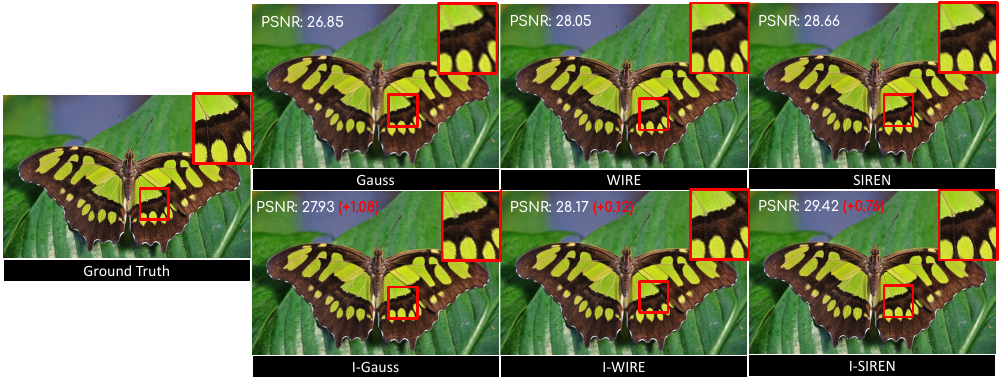}
    \caption{Visual quality comparison of super-resolution results at 2$\times$ scale using various methods. The ground truth is compared against baseline INR methods SIREN, WIRE, Gauss, and their iterative counterparts. The iterative approaches consistently achieve sharper reconstructions with fewer artifacts, effectively preserving finer details and high-frequency structures.}
    \label{fig:SR}
\end{figure*}

\noindent
\paragraph{Network Architecture.} 
Figure~\ref{fig:architecture} provides an overview of our plug-and-play I-INR framework, which comprises three main components: a Backbone network, a FeedbackNet module, and a FuseNet module. The Backbone extracts initial features from the input; the FeedbackNet incorporates the intermediate state and time conditioning; and the FuseNet merges these feature streams. The final output is given by Eq.~(\ref{eq:fusenet}). This design enables flexible integration with existing INR architectures and supports efficient iterative refinement.
The information flow between components is formalized as
\begin{equation}
f_\theta\bigl(\hat{g}(x)_t,\,x,\,t\bigr)
  = \operatorname{FuseNet}\bigl(\operatorname{concat}(\mathbf{f},\mathbf{b})\bigr) \odot \mathbf{b},
\label{eq:fusenet}
\end{equation}
where
\[
\mathbf{b} := \operatorname{Backbone}(x), \qquad
\mathbf{f} := \operatorname{FeedbackNet}\bigl(\hat{g}(x)_t,\,x,\,t\bigr).
\]
At each iteration, the current state is processed by the FeedbackNet, and its output is fused with the Backbone features via FuseNet. Notably, the Backbone is forward-passed only once throughout the reconstruction, while the lightweight FeedbackNet and FuseNet operate at each iteration, resulting in a substantial reduction in computational cost.

\section{Experiments}
\label{exp}
\ali{
To evaluate the effectiveness and robustness of our proposed method, we \tahir{conduct} extensive experiments across diverse tasks and backbones. Our evaluation encompasses regression tasks, such as 2D and 3D signal fitting, as well as generalization and robustness tasks, including image SR and image denoising. To demonstrate its comparative advantage, we \tahir{benchmark} our method against established baseline approaches, including SIREN, Gauss, and WIRE.
}

\noindent
\paragraph{Implementation Settings. }
\ali{The proposed I-INR framework is implemented in PyTorch~\cite{paszke2019pytorch} and optimized using the Adam optimizer~\cite{diederik2014adam}. For all experiments, the noise parameter $\epsilon$ \tahir{is} empirically set to 0.1, and the inference $steps$ \tahir{are} fixed to 2 unless stated otherwise. The initial state $\mathcal{Z}$ is sampled from a standard normal Gaussian distribution across all experiments. All baseline methods, including SIREN, WIRE, and Gauss, as well as their iterative counterparts, \tahir{are} initialized following the configurations specified in their respective papers. In our implementation, both FeedbackNet and FuseNet are lightweight MLPs with two layers, each layer having a width of 30 for FeedbackNet and 100 for FuseNet.}





\noindent
\paragraph{Evaluation Metrics.}
\salman{To comprehensively evaluate our proposed method, we conduct extensive experiments using both distortion-based metrics, such as Peak Signal-to-Noise Ratio (PSNR) and Structural Similarity Index Measure (SSIM)~\cite{wang2004image}, as well as perceptual metrics like Learned Perceptual Image Patch Similarity (LPIPS)~\cite{zhang2018unreasonable}. For a detailed analysis, refer to the supplementary materials.}
\subsection{Results}
\salman{Our results consistently demonstrate that I-INR outperforms existing methods across diverse tasks, including image fitting, SR, and 3D occupancy. It achieves superior reconstruction quality and enhanced robustness compared to baseline approaches (SIREN, WIRE, Gauss).}


\begin{figure*}[t]
    \centering
    \includegraphics[width=\linewidth]{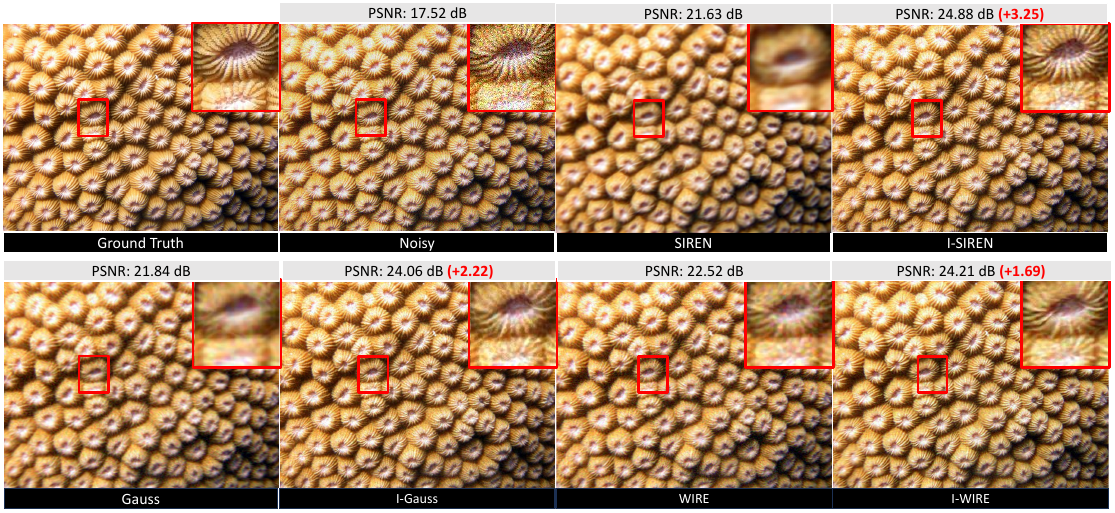}
    \caption{
    Visual comparison of denoising results using various methods. The ground truth is compared against noisy input, baseline methods SIREN, WIRE, and Gauss, as well as their iterative counterparts. I-INR demonstrates superior artifact reduction and detail preservation compared to their non-iterative counterparts.}
    \label{fig:denoising}

\end{figure*}

\begin{table}[t]
    \centering

    \small
    \setlength{\tabcolsep}{1.5pt}
    \renewcommand{\arraystretch}{0.9}
    \begin{tabular}{l|cc|cc|cc}
    \toprule
        & \multicolumn{2}{c|}{\textbf{SIREN}} & \multicolumn{2}{c|}{\textbf{WIRE}} & \multicolumn{2}{c}{\textbf{Gauss}} \\
        & PSNR~$\uparrow$ & SSIM~$\uparrow$ & PSNR~$\uparrow$ & SSIM~$\uparrow$ & PSNR~$\uparrow$ & SSIM~$\uparrow$ \\
    \midrule
        Baseline  & 34.57 & 0.931 & 32.15 & 0.898 & 31.33 & 0.880 \\
        Ours      & \textbf{37.53} & \textbf{0.961} & \textbf{33.73} & \textbf{0.924} & \textbf{31.93} & \textbf{0.884} \\
    \bottomrule
    \end{tabular}
    \caption{
    Image fitting results on Kodak comparing SIREN, WIRE, and Gauss with their iterative counterparts.}
    \label{tab:image_fitting}

\end{table}


\noindent
\paragraph{Image Fitting.}
\ali{For image fitting, each baseline employs a 3-layer MLP with 300 neurons per layer, serving as the Backbone INR for iterative models (Figure~\ref{fig:architecture}). All networks are trained on the full-resolution Kodak dataset~\cite{kodak}, with average results reported in Table~\ref{tab:image_fitting}. The proposed iterative models consistently outperform their one-shot counterparts. As shown in Figure~\ref{fig:Image_Fitting}, error maps from baseline models display uniform noise, especially in high-frequency regions, whereas iterative variants produce smoother reconstructions with reduced errors, including in challenging areas.}

\noindent
\paragraph{Image Super-resolution.}
\ali{We evaluate our I-INR framework against traditional one-shot INRs on SR tasks using 40 images from the DIV2K dataset \cite{agustsson2017ntire}. The model \tahir{is} trained exclusively on $2\times$ SR with a two-layer MLP of 256 neurons per layer.
} \ali{To assess generalization, we evaluate I-INR on both $2\times$ and $4\times$ SR, despite training only on $2\times$. Table~\ref{tab:SR} compares its performance with single-shot baselines, demonstrating that our approach consistently enhances fidelity (PSNR) and perceptual quality (LPIPS) across all architectures.} \ali{Figure~\ref{fig:SR} visualizes the butterfly image for $2\times$ SR (refer to supplementary for visualization at $4\times$), comparing each baseline with its iterative counterpart. The proposed approach consistently outperforms baseline methods, generating sharper reconstructions with fewer artifacts across all non-linearities.}

\begin{table}[t]
    \centering
    \small
    \setlength{\tabcolsep}{1.5pt}
    \renewcommand{\arraystretch}{0.9}
    \resizebox{\columnwidth}{!}{
    \begin{tabular}{c|l|cc|cc|cc}
    \toprule
        \textbf{Scale} & \textbf{Method} & \multicolumn{2}{c|}{\textbf{SIREN}} & \multicolumn{2}{c|}{\textbf{WIRE}} & \multicolumn{2}{c}{\textbf{Gauss}} \\
        & & PSNR~$\uparrow$ & LPIPS~$\downarrow$ & PSNR~$\uparrow$ & LPIPS~$\downarrow$ & PSNR~$\uparrow$ & LPIPS~$\downarrow$ \\
    \midrule
        \multirow{2}{*}{$2\times$} & Baseline & 26.77 & 0.414 & 26.14 & 0.457 & 25.19 & 0.538 \\
                                   & Ours     & \textbf{27.64} & \textbf{0.367} & \textbf{27.21} & \textbf{0.388} & \textbf{26.82} & \textbf{0.363} \\
    \midrule
        \multirow{2}{*}{$4\times$} & Baseline & 25.03 & 0.597 & 24.57 & 0.618 & 23.85 & 0.673 \\
                                   & Ours     & \textbf{25.53} & \textbf{0.575} & \textbf{25.78} & \textbf{0.496} & \textbf{25.18} & \textbf{0.620} \\
    \bottomrule
    \end{tabular}
    }
    \caption{
    Comparison of (SR) performance for $2\times$ and $4\times$ upscaling using a model trained on $2\times$ SR. Results are shown alongside SIREN, WIRE, and Gauss, including their iterative counterparts. The best results are shown in bold.}
    \label{tab:SR}
\end{table}

\begin{table}[t] 
    \centering

    \small
    \setlength{\tabcolsep}{1.5pt}
    \renewcommand{\arraystretch}{0.9}
    \begin{tabular}{l|cc|cc|cc}
    \toprule
        & \multicolumn{2}{c|}{\textbf{SIREN}} & \multicolumn{2}{c|}{\textbf{WIRE}} & \multicolumn{2}{c}{\textbf{Gauss}} \\
        & PSNR~$\uparrow$ & LPIPS~$\downarrow$ & PSNR~$\uparrow$ & LPIPS~$\downarrow$ & PSNR~$\uparrow$ & LPIPS~$\downarrow$ \\
    \midrule
        Baseline  & 23.86 & 0.604 & 23.32 & 0.746 & 23.10 & 0.783 \\
        Ours      & \textbf{25.59} & \textbf{0.540} & \textbf{24.76} & \textbf{0.490} & \textbf{24.20} & \textbf{0.533} \\
    \bottomrule
    \end{tabular}
    \caption{
    Image denoising results on 40 images sampled from the DIV2K dataset, comparing SIREN, WIRE, and Gauss with their iterative counterparts.}
    \label{tab:denoising}
    
\end{table}


\begin{figure*}[t]
    \centering
    \includegraphics[width=0.85\linewidth]{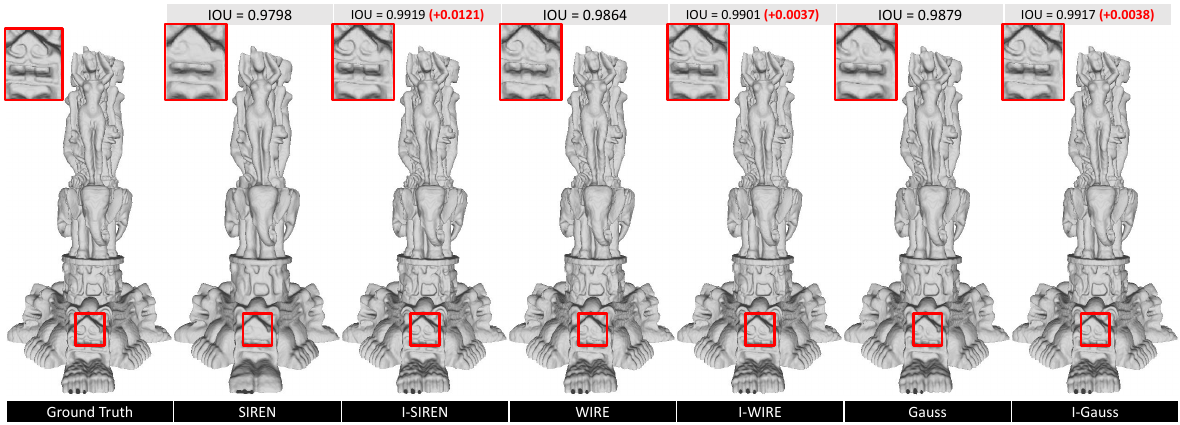}
    \caption{
    Visualization of 3D occupancy reconstruction results using various methods and their iterative counterparts. The ground truth is compared to baseline models SIREN, WIRE, Gauss, and their iterative counterparts.} 
    \label{fig:img_occpancy}
\end{figure*}

\noindent
\paragraph{Image Denoising.}
To assess the robustness of I-INRs in modeling noisy signals, we utilize 40 high-resolution color images from the DIV2K dataset and introduce Poisson-distributed noise to each pixel, with a maximum mean photon count of 30 and a readout noise level of 2, following ~\cite{saragadam2023wire}. The same architecture and training procedure as SR is employed. 
\ali{Table~\ref{tab:denoising} presents quantitative results comparing the denoising performance of baseline architectures with their iterative counterparts. 
Iterative models consistently achieve significant improvements in both PSNR and LPIPS over their single-shot baselines.}
\ali{Figure~\ref{fig:denoising} provides a qualitative comparison of denoising performance for SIREN, WIRE, and Gauss alongside their iterative versions. The iterative approach significantly enhances reconstruction quality across all architectures, achieving up to +3.25 dB PSNR improvement (for SIREN).}

\begin{figure*}[t]
    \centering
    \includegraphics[width=\linewidth]{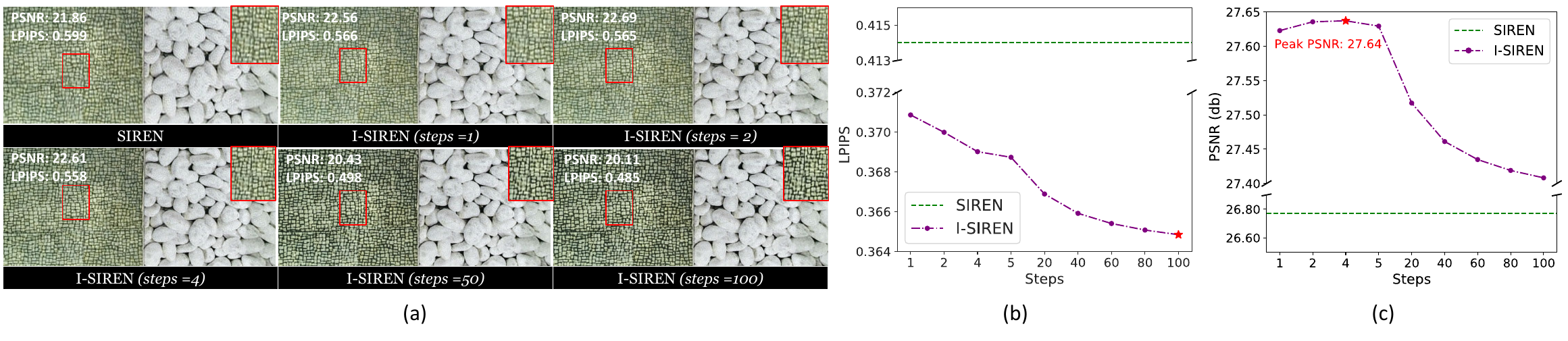}
    \caption{Analysis of I-SIREN's performance for super-resolution at 2$\times$ resolution across different iteration $steps$. (a) I-SIREN achieves peak PSNR at $steps=4$, where further increasing time steps enhances perceptual quality but reduces fidelity due to the fidelity-perception tradeoff. (b) Increasing the number of steps consistently improves perceptual quality. (c) PSNR reaches its highest performance at $steps=4$, and while additional iterations slightly reduce PSNR, it remains superior to the baseline.}
    \label{fig:steps_analysis}
\end{figure*}

\noindent
\paragraph{3D Occupancy.}
\salman{The 3D occupancy task aims to reconstruct 3D shapes by modeling the occupancy of points in space. We conduct experiments on the Armadillo, Dragon, and Thai Statue 3D scenes~\cite{gal2006salient}, utilizing a two-layer MLP with 256 neurons per layer for both the baseline INRs and their iterative counterparts.  Table~\ref{tab:iou_results} presents the averaged Intersection over Union (IoU) results, while Figure~\ref{fig:img_occpancy} visualizes the reconstructions achieved using different nonlinearities and their iterative variants. Notably, I-INRs outperform their single-shot baselines, generating sharper and more precise reconstructions. The iterative approach enhances detail preservation, particularly in complex regions as shown in Figure~\ref{fig:img_occpancy}. 
}
\begin{table}[t]
    \centering
    \renewcommand{\arraystretch}{0.9}
    \begin{tabular}{l|c|c|c}
    \toprule
        & \textbf{SIREN}~$\uparrow$ & \textbf{WIRE}~$\uparrow$ & \textbf{Gauss}~$\uparrow$ \\
    \midrule
        Baseline  & 0.9840 & 0.9917 & 0.9855 \\
        Ours      & \textbf{0.9934} & \textbf{0.9950} & \textbf{0.9967} \\
    \bottomrule
    \end{tabular}
    \caption{
    Averaged Intersection over Union (IoU) results comparing SIREN, WIRE, and Gauss with their iterative counterparts on the Armadillo, Dragon, and Thai Statue 3D.}
    \label{tab:iou_results}
\end{table}


\subsection{Ablation Studies}
\salmaan{We ablate inference complexity, Feedback/FuseNet impact, and reconstruction steps. Further analysis on initial state, training cost, layer depth, and statistical significance is included in the supplementary material.}
\noindent


\noindent
\paragraph{Inference Complexity.}
\salmaan{During reconstruction, I-SIREN executes the Backbone network only once, while subsequent refinement steps involve forward passes through the lightweight FeedbackNet and FuseNet modules. The iterative setup introduces minimal computational overhead: the Backbone accounts for 106.8 GFLOPs, and each refinement step adds just 0.43 GFLOPs. For two refinement steps ($\text{steps} = 2$), the total computational cost amounts to 107.6 GFLOPs, approximately equivalent to the original SIREN.}
\noindent
\salmaan{\paragraph{Impact of FeedbackNet and FuseNet:}
To evaluate the impact of architectural components of I-INR, we performed ablation studies by selectively disabling the FeedbackNet and FuseNet modules and varying their parameter scales. These experiments were conducted on the image fitting task using I-SIREN on the Kodak dataset. As summarized in Table~\ref{table:ablation_kodak}, removing both modules leads to a significant performance drop, with PSNR falling to 20.27. Introducing FeedbackNet alone substantially improves reconstruction quality, and the addition of FuseNet provides further gains. The best performance is achieved when both modules are included, with moderate parameter scaling yielding an optimal balance between computational cost and accuracy.} 

\begin{table}[t]
    \centering
    \small
    \setlength{\tabcolsep}{3pt}

    \begin{tabular}{c|c|c}
        \toprule
        \textbf{FeedbackNet (Params)} & \textbf{FuseNet (Params)} & \textbf{PSNR} \\
        \midrule
        \xmark & \xmark & 20.27 \\
        \cmark~(1$\times$) & \xmark & 31.88 \\
        \cmark~(1$\times$) & \cmark~(1$\times$) & 37.53 \\
        \cmark~(2$\times$) & \cmark~(1$\times$) & 37.25 \\
        \cmark~(1$\times$) & \cmark~(2$\times$) & 37.77 \\
        \cmark~(2$\times$) & \cmark~(2$\times$) & 37.56 \\
        \bottomrule
    \end{tabular}
    \caption{Impact of FeedbackNet and FuseNet on image fitting task with I-SIREN on the Kodak dataset.}
    \label{table:ablation_kodak}
\end{table}

\noindent
\paragraph{Analysis of Reconstruction Steps. }
\salmaan{We evaluate the effect of reconstruction steps on $2{\times}$ super-resolution using 40 images from the DIV2K dataset (Figures~\ref{fig:steps_analysis}b and \ref{fig:steps_analysis}c). PSNR peaks at $steps=4$, with the largest improvement from 1 to 2 steps; beyond 2, gains saturate. While PSNR slightly drops for $steps>4$, perceptual quality (LPIPS) continues to improve, reflecting the Perception-Distortion Tradeoff~\cite{blau2018perception}. Figure~\ref{fig:steps_analysis}a visually compares SIREN and I-SIREN across steps, showing that I-SIREN produces increasingly vivid and detailed reconstructions, confirming that additional refinement steps are beneficial for capturing finer textures and high-frequency components. This trend generalizes to other tasks, as discussed in the supplementary material.}
\section{Conclusion}
\salman{This work introduces a novel I-INR framework that progressively reconstructs a signal. The proposed method serves as a plug-and-play solution compatible with existing INR methods, significantly enhancing their performance. Through extensive experiments, we demonstrate that I-INR effectively captures high-frequency details, improves noise robustness, and achieves superior reconstruction quality across various tasks and nonlinearities.}


{
    \small
    \bibliography{main}
}

\setcounter{page}{1}
\setcounter{figure}{7}  
\setcounter{table}{5}  
\maketitle 

\section{Additional Results}
\subsection{Additional Quantative Results}
Table~\ref{supp-tab:results} presents a detailed comparison of our proposed method against the baseline across image fitting, denoising, and SR at $2\times$ for all three evaluation metrics (PSNR, SSIM, and LPIPS). Our iterative approach consistently enhances performance across tasks and different baselines, achieving superior reconstruction quality.  

\subsection{Additional Qualitative Comparison}
\begin{figure*}[t]
    \centering
    \includegraphics[width=\linewidth]{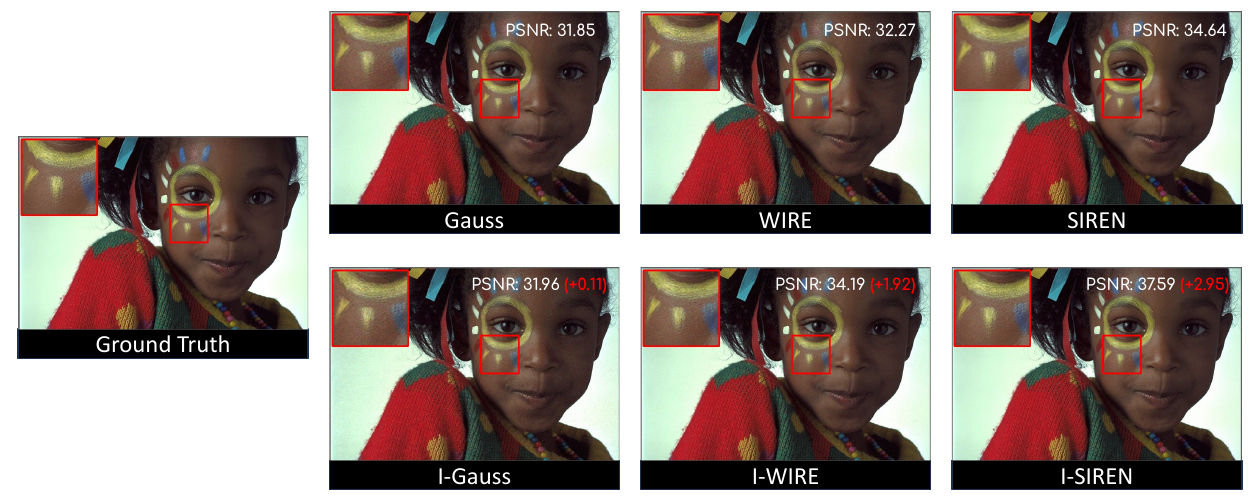}
    \caption{Image fitting results for different non-linearities and their iterative extensions (prefix "I").}
    \label{supp_fig:Image_Fitting}
\end{figure*}

\begin{figure*}[t]
    \centering
    \includegraphics[width=\linewidth]{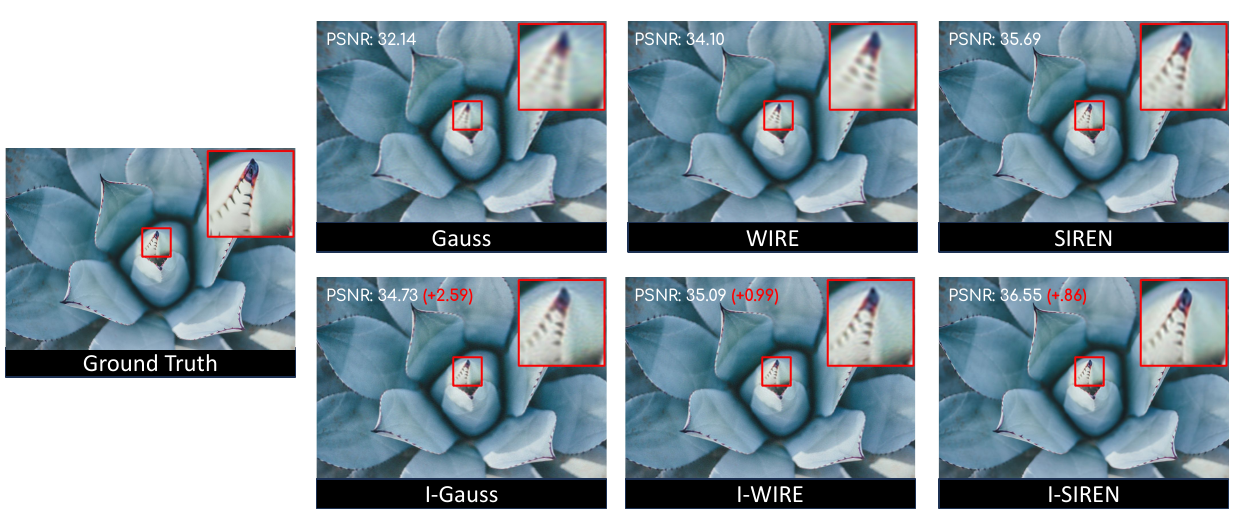}
    \caption{Visual quality comparison of super-resolution results at 2$\times$ scale using various methods. The ground truth is compared against baseline INR methods SIREN, WIRE, Gauss, and their iterative counterparts. The iterative approaches consistently achieve sharper reconstructions with fewer artifacts, effectively preserving finer details and high-frequency structures.}
    \label{supp_fig:SR}
\end{figure*}

\begin{figure*}[t]
    \centering
    \includegraphics[width=\linewidth]{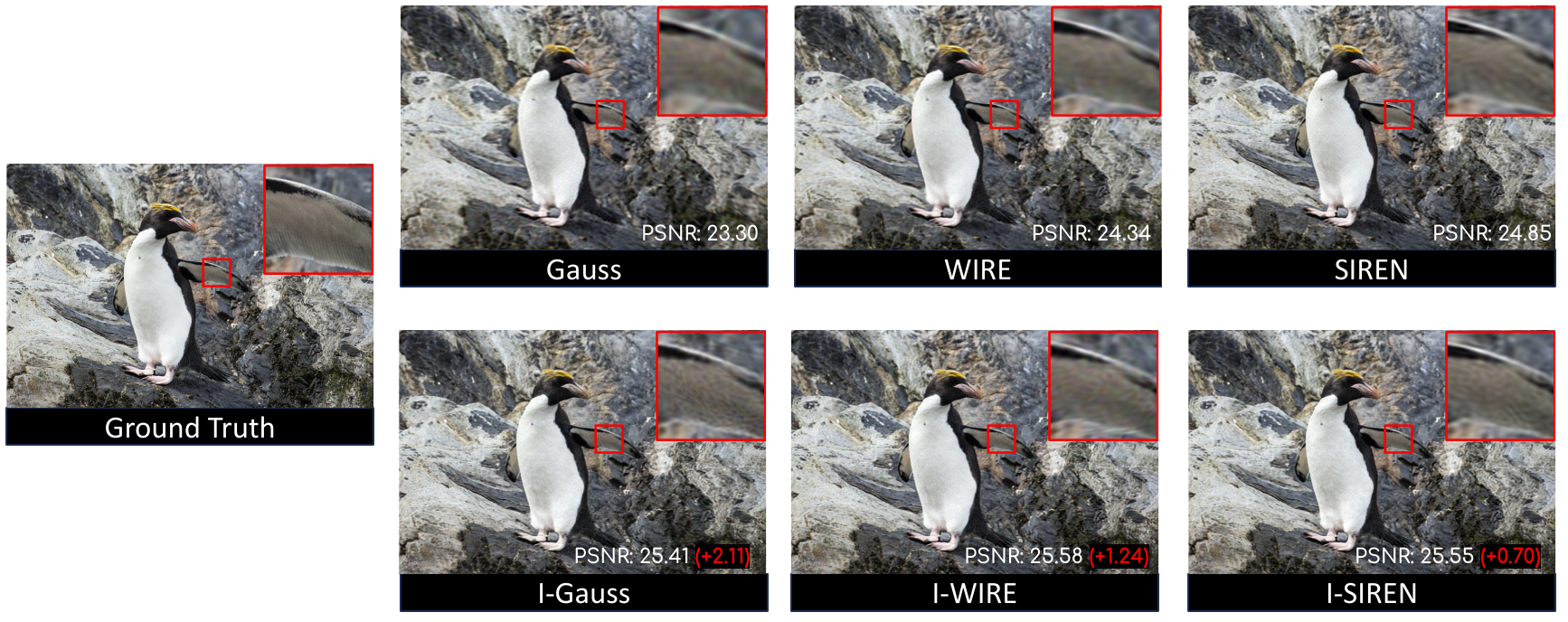}
    \caption{Visual quality comparison of super-resolution results at 4$\times$ scale using various methods. The ground truth is compared against baseline INR methods SIREN, WIRE, Gauss, and their iterative counterparts. The iterative approaches consistently achieve sharper reconstructions with fewer artifacts, effectively preserving finer details and high-frequency structures.}
    \label{supp_fig:SR-4x}
\end{figure*}

\begin{figure*}[t]
    \centering
    \includegraphics[width=\linewidth]{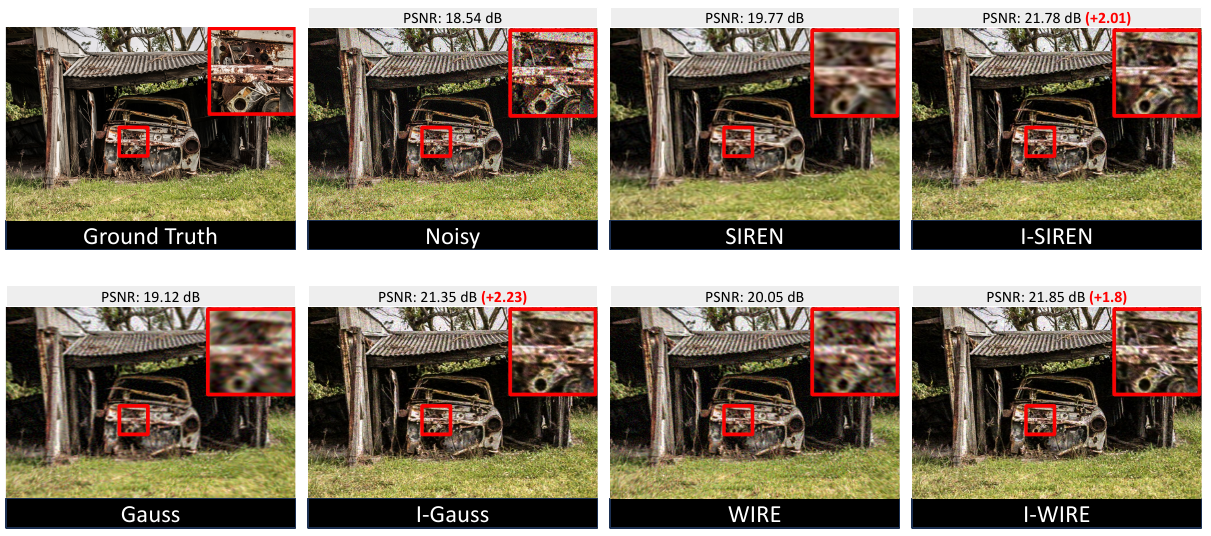}
    \caption{Visual comparison of denoising results using various methods. The ground truth is compared against noisy input, baseline methods SIREN, WIRE, and Gauss, as well as their iterative counterparts.
    I-INR demonstrates superior artifact reduction and detail preservation compared to their non-iterative counterparts.}
    \label{supp_fig:denoising}
\end{figure*}

We also present an additional qualitative comparison of our proposed method against SIREN, Gauss, and WIRE. Figures~\ref{supp_fig:Image_Fitting},~\ref{supp_fig:SR},~\ref{supp_fig:SR-4x},~\ref{supp_fig:denoising}, illustrates a comparative analysis for image fitting, SR at $2\times$ and $4\times$ scales, and image denoising, demonstrating the improvements achieved by our approach over baseline INRs. Our proposed method consistently outperforms single-shot baselines, demonstrating superior reconstruction quality across various models and tasks.

\section{Additional Ablation Studies}

\paragraph{Initial State.}
\ali{I-INR reconstructs a signal iteratively over multiple steps, starting from an initial state \( \mathcal{Z} \). The choice of this initial state can significantly impact the reconstruction process. To assess its effect, we \tahir{conduct} extensive experiments with different types of initial states. Empirically, \( \mathcal{Z} \) sampled from a standard normal Gaussian distribution $\mathcal{N}(0,1)$ consistently outperforms all-zeros and all-ones initial states across different non-linearities for the image fitting task. The results, summarized in Table~\ref{tab:initialize}, highlight the superiority of noise-based initial states in achieving higher PSNR and SSIM values.}

\begin{figure*}
    \begin{tabular}{@{}c@{}c@{}}
        \includegraphics[width=0.5\linewidth]{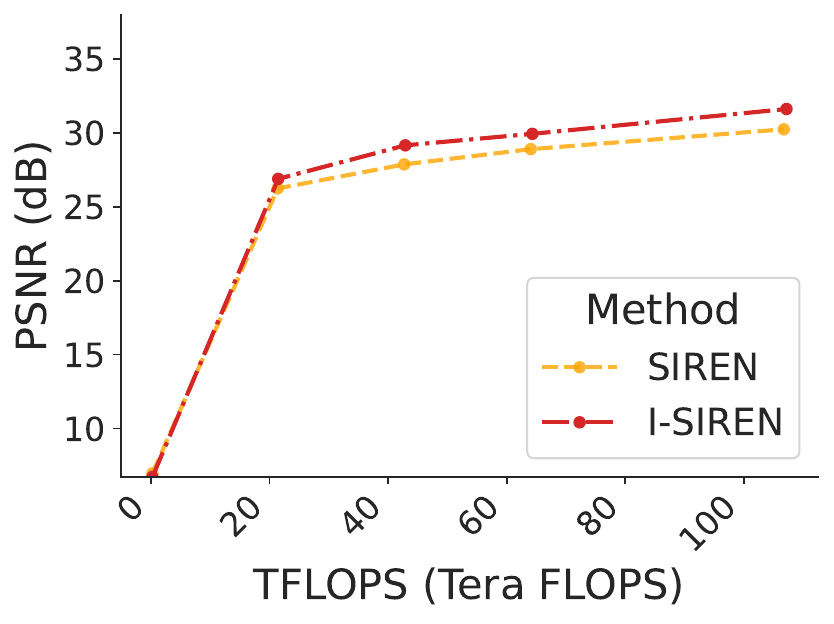} &
        \includegraphics[width=0.5\linewidth]{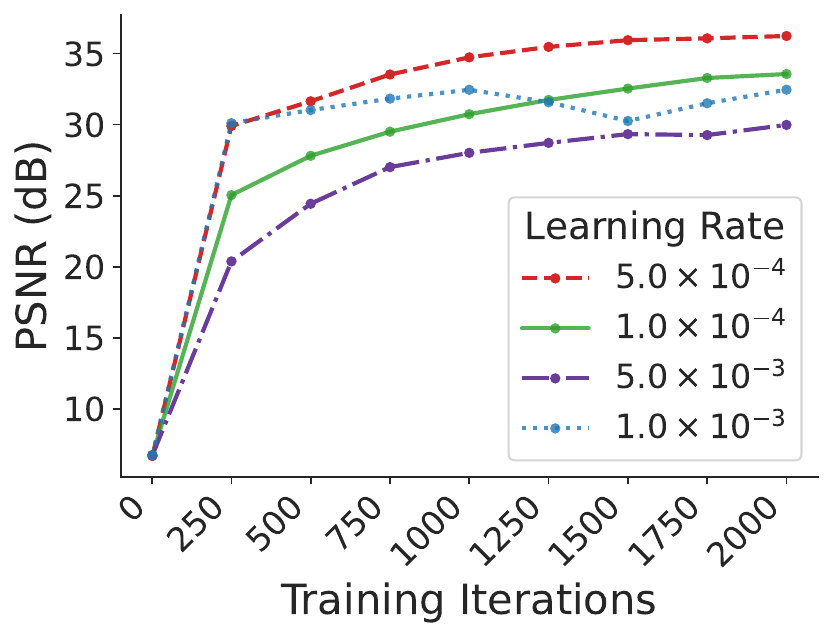} \\ 
        (a)&(b)\\
    \end{tabular}
    \caption{On the Kodak dataset for image fitting: (a) PSNR vs. training FLOPs comparison between SIREN and I-SIREN. (b) PSNR vs. training iterations for I-SIREN with different learning rates.}
    \label{fig:opt_det}
\end{figure*}

\noindent
\textbf{Training Complexity.}
\salman{Our proposed I-INR method achieves faster convergence than the non-iterative baseline while maintaining similar training complexity, as shown in Figure~\ref{fig:opt_det}a. The results are averaged over the Kodak dataset for the image fitting task. Figure~\ref{fig:opt_det}b further illustrates I-SIREN’s training performance under different learning rates.} 

\noindent
\textbf{Number of Layers in Backbone Network.}
\ali{For the image fitting task on the Kodak dataset, we investigate the impact of increasing the number of layers in SIREN and its iterative counterpart, I-SIREN. The results, presented in Table~\ref{table:layers}, show that while the performance of SIREN saturates beyond a certain depth, specifically at five layers, I-SIREN continues to improve as more layers are added. Notably, a three-layer I-SIREN outperforms a five-layer SIREN, demonstrating the efficiency of the iterative approach.}

\begin{figure*}[t]
    \centering
    \includegraphics[width=\linewidth]{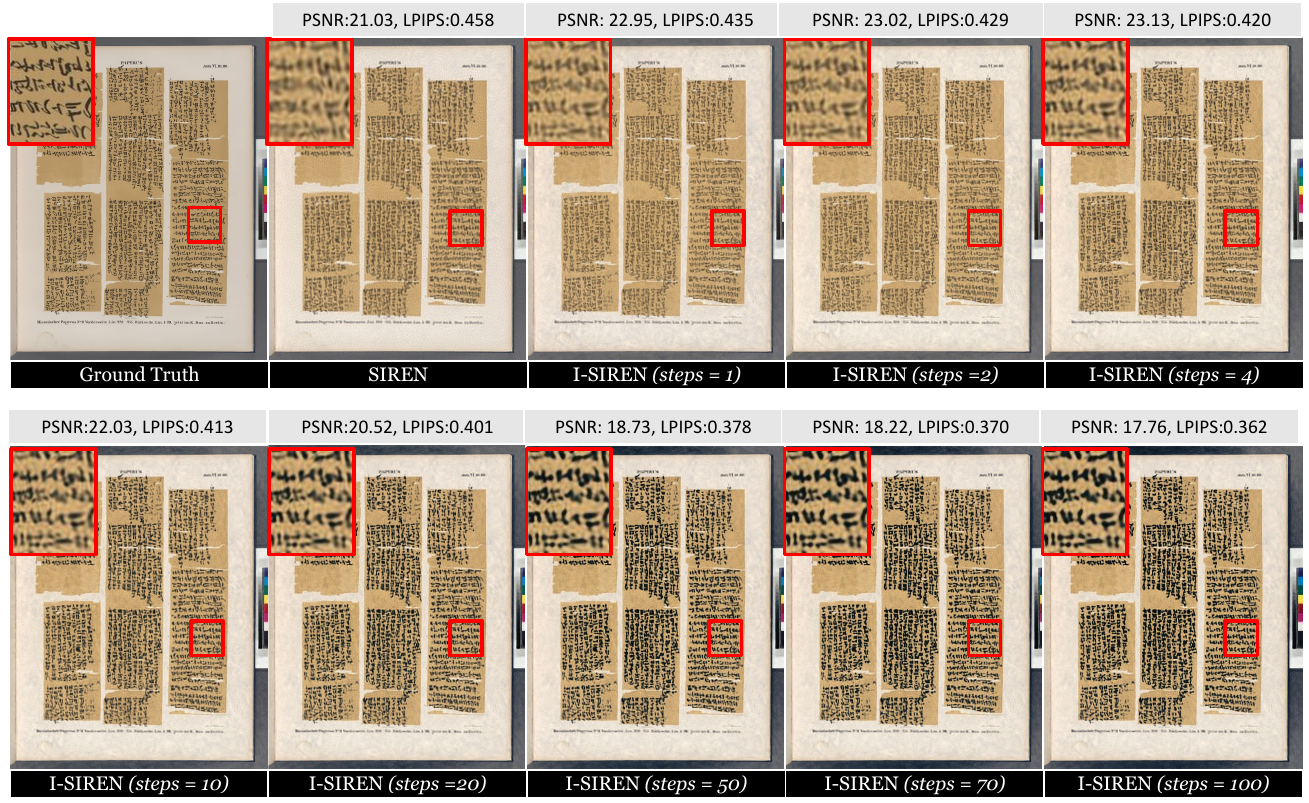}
    \caption{Analysis of I-SIREN's performance for super-resolution at 2$\times$ resolution across different steps. I-SIREN achieves peak PSNR at $steps=4$, where further increasing steps enhances perceptual quality.}
    \label{supp-fig:steps_analysis}
\end{figure*}

\noindent
\textbf{Additional Reconstruction Steps Analysis.} For SR at $2\times$, we further analyze the reconstruction quality of I-SIREN over multiple steps, comparing it to SIREN and the ground truth. As shown in Figure~\ref{supp-fig:steps_analysis}, the results demonstrate that with an increasing number of steps, details become more pronounced, and more high-frequency information is incorporated into the reconstruction.

\begin{figure*}
    \begin{tabular}{@{}c@{}c@{}}
        \includegraphics[width=0.5\linewidth]{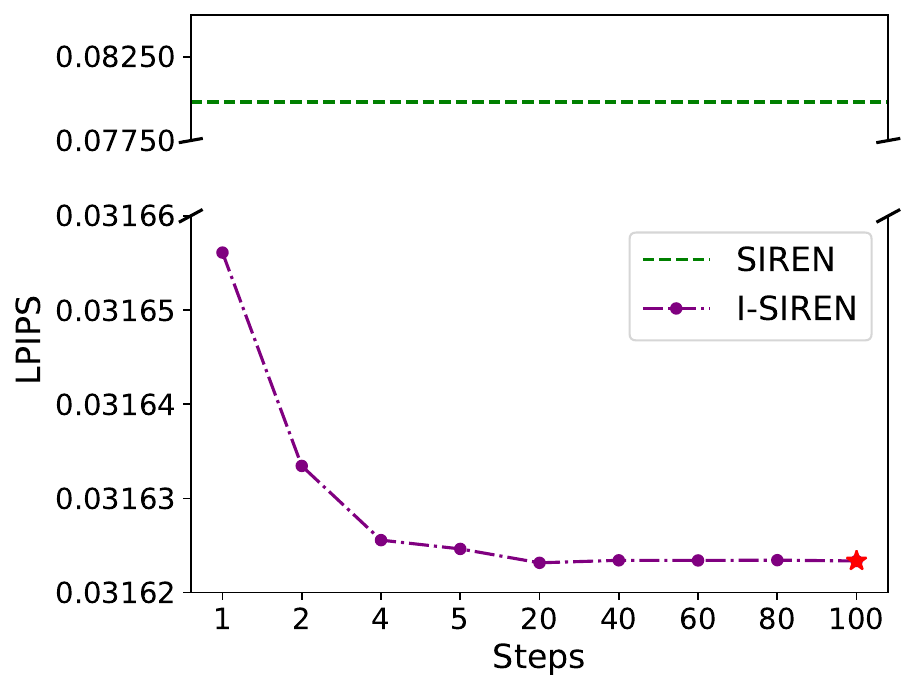} &
        \includegraphics[width=0.5\linewidth]{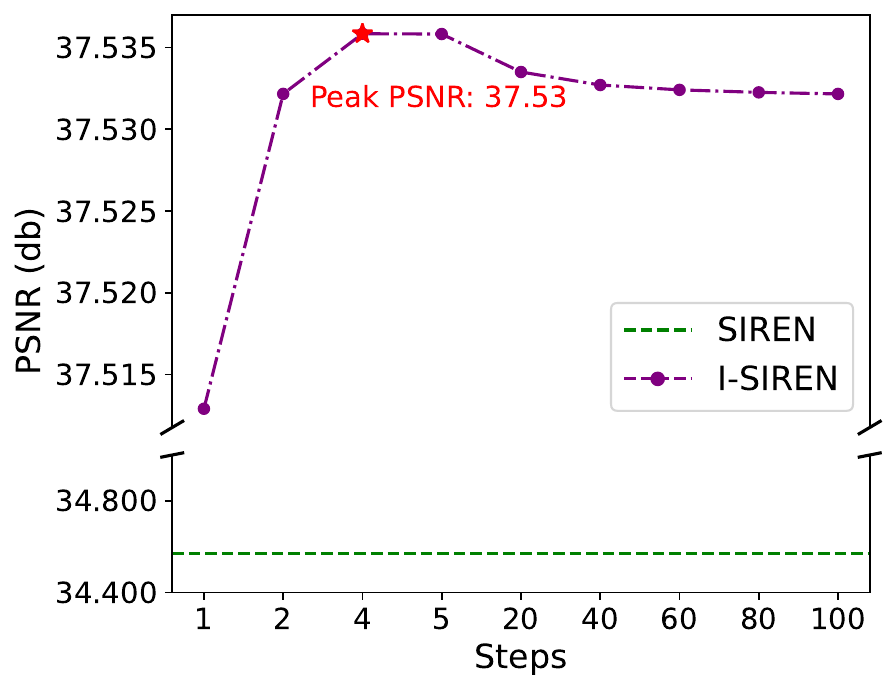} \\ 
        (a)&(b)\\
    \end{tabular}
    \vspace{-10pt}
    \caption{Analysis of I-SIREN's performance for image fitting across different iteration steps.  (a) Increasing the number of steps consistently improves perceptual quality. (b) PSNR reaches its highest performance at $steps=4$, and while additional iterations slightly reduce PSNR, it remains superior to the baseline.}
    \label{supp-fig:opt_det}
\end{figure*}

\noindent
Additionally, We analyze the impact of reconstruction steps on image fitting using the Kodak dataset (see Figure~\ref{supp-fig:opt_det}). Our findings indicate that the model achieves its highest PSNR at \(steps=4\), with the most significant gain occurring between \(steps=1\) and \(steps=2\). Beyond \(steps=2\), PSNR begins to plateau. Notably, even though PSNR decreases for \(steps>4\), perceptual quality continues to improve.

\begin{figure*}
    \begin{tabular}{@{}c@{}c@{}}
        \includegraphics[width=0.5\linewidth]{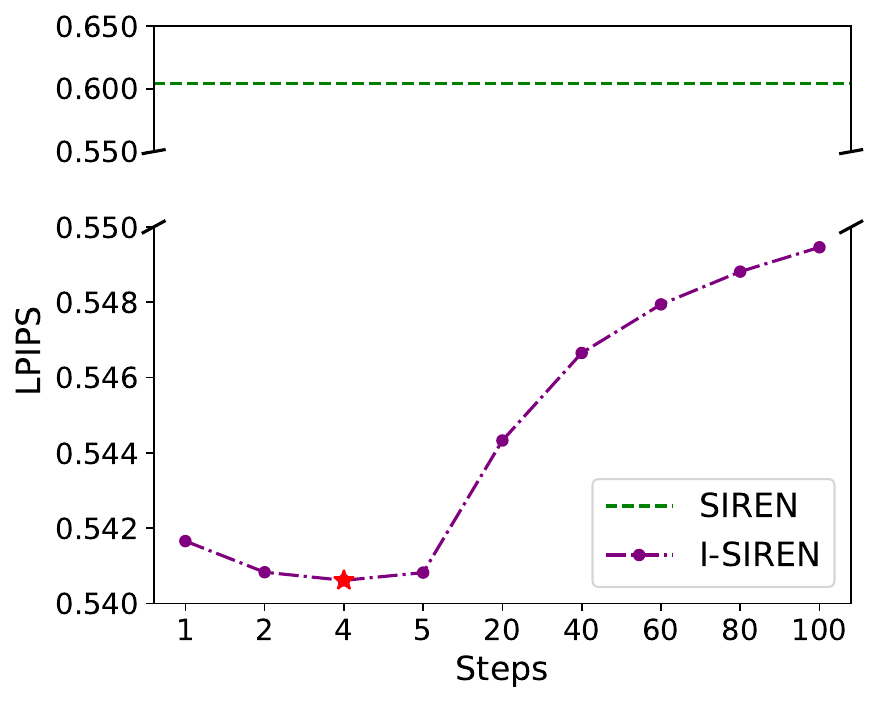} &
        \includegraphics[width=0.5\linewidth]{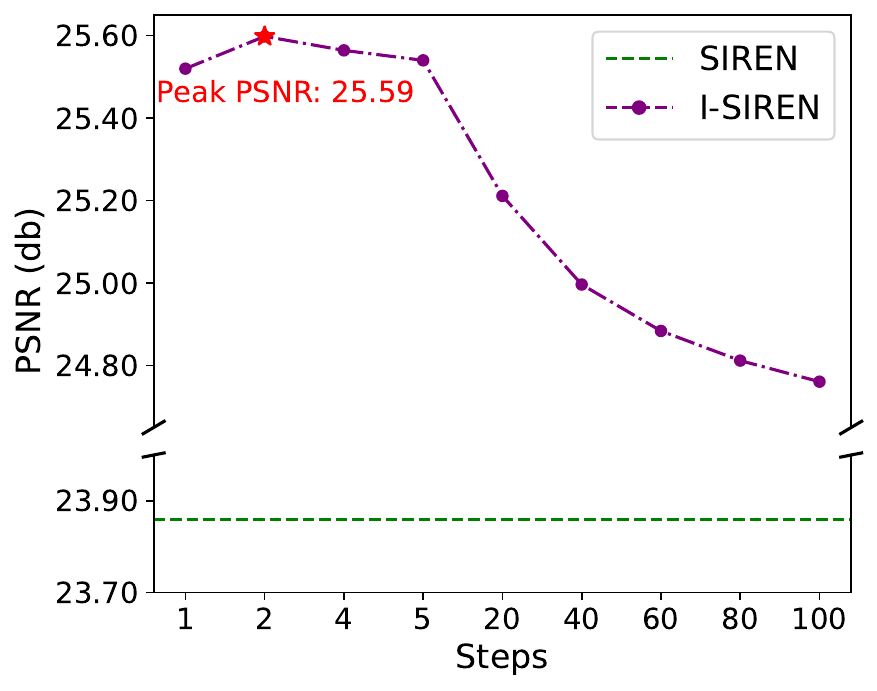} \\ 
        (a)&(b)\\
    \end{tabular}
    \vspace{-10pt}
    \caption{Analysis of I-SIREN’s performance for denoising across different reconstruction steps. (a) LPIPS improves until step 4 and stabilizes around step 5. (b) PSNR reaches its peak at \(steps=2\), but further iterations degrade performance.}
    \label{supp-fig:opt_den}
\end{figure*}

We also analyze the effect of reconstruction steps on image denoising performance using 40 images from the DIV2K dataset (see Figure~\ref{supp-fig:opt_den}). Our findings indicate that PSNR peaks at $steps=2$ before gradually declining as $steps$ increases. Meanwhile, LPIPS improves until $step=4$ and stabilizes around step 5. Beyond $steps=4$, both PSNR and LPIPS degrade as I-INR begins reconstructing noise present in the data. At higher iterations, the model inadvertently reconstructs these distortions, leading to a decline in performance.  

\noindent
\noindent
\textbf{Feature Fusion.}
We conduct ablation studies on the final feature-fusion mechanism for signal reconstruction, evaluating two strategies that combine the outputs of the \emph{Backbone} and \emph{FuseNet} modules (Figure~2a of main manuscript):
\[
\mathbf{b} := \operatorname{Backbone}(x), \qquad
\mathbf{f} := \operatorname{FeedbackNet}(\hat{g}(x)_t,\, x,\, t)
\]
\[
\mathbf{z} := \operatorname{FuseNet}(\operatorname{concat}(\mathbf{f},\, \mathbf{b}))
\]
\[
\text{Multiplicative:} \quad \mathit{out} = \mathbf{b} \odot \mathbf{z},
\]
\[
\text{Adaptive:} \quad \mathit{out} = \mathbf{b} \cdot t + \mathbf{z} \cdot (1 - t)
\]

In \emph{Multiplicative} fusion, the final output is obtained via element-wise multiplication of the Backbone and FuseNet outputs. The \emph{Adaptive} method, conversely, uses the interpolation factor $t$ to smoothly shift emphasis from the Backbone output (dominant at $t=1$) toward the FuseNet output at higher values of $t$. Quantitative results comparing both fusion strategies for the image-fitting task are summarized in Table~\ref{tab:fusion_comparison_step2}. Multiplicative fusion consistently outperforms Adaptive fusion, as evidenced by the results. Therefore, we adopt Multiplicative fusion as the standard approach for our method.

\noindent
\textbf{Time Complexity.}
We compare SIREN and I-SIREN on the Kodak dataset, trained for 2000 iterations on an NVIDIA RTX\,4090 GPU. 
As shown in Table~\ref{tab:time_comparison}, I-SIREN adds only a small overhead in training time ($+6\%$) and inference latency ($+3$\,ms per image) while achieving higher reconstruction quality, confirming its efficiency.

\noindent
\textbf{Statistical Significance.}
All prior experiments were conducted using a single random seed. To assess the robustness and statistical significance of our approach, we additionally evaluate performance across five different seeds. Table~\ref{tab:std} reports the results for image fitting, SR, and denoising tasks, comparing SIREN and I-SIREN, along with confidence intervals. The dataset and experimental settings remain consistent with those previously described. As shown in Table~\ref{tab:std}, I-SIREN consistently outperforms SIREN across all tasks. The improvements achieved by I-SIREN are statistically significant, with \textit{p}-values $<$ 0.00001 based on the Wilcoxon signed-rank test.

\begin{table*}[t]
    \centering
    \begin{tabular}{ll|l|ccc|ccc|ccc}
        \toprule
        & & & \multicolumn{3}{c|}{\textbf{SIREN}} & \multicolumn{3}{c|}{\textbf{Gauss}} & \multicolumn{3}{c}{\textbf{WIRE}} \\
        & Task & Model & PSNR$\uparrow$ & SSIM$\uparrow$ & LPIPS$\downarrow$ & PSNR$\uparrow$ & SSIM$\uparrow$ & LPIPS$\downarrow$ & PSNR$\uparrow$ & SSIM$\uparrow$ & LPIPS$\downarrow$ \\
        \midrule        
        \multirow{2}{*} & Image Fitting & Baseline & 34.57 & 0.931 & 0.080 & 31.33 & 0.880 & 0.147 & 32.15 & 0.898 & 0.120 \\
        & & Ours & \textbf{37.53} & \textbf{0.961} & \textbf{0.032} & \textbf{31.93} & \textbf{0.884} & \textbf{0.124} & \textbf{33.73} & \textbf{0.924} & \textbf{0.077} \\
        \midrule
        \multirow{2}{*} & Super Resolution & Baseline & 26.77 & 0.763 & 0.414 & 25.19 & 0.709 & 0.538 & 26.14 & 0.749 & 0.457 \\
        & & Ours & \textbf{27.64} & \textbf{0.783} & \textbf{0.367} & \textbf{26.82} & \textbf{0.761} & \textbf{0.363} & \textbf{27.21} & \textbf{0.778} & \textbf{0.388} \\
        \midrule
        \multirow{2}{*} & Denoising & Baseline & 23.86 & \textbf{0.658} & 0.604 & 23.09 & 0.604 & 0.783 & 23.32 & 0.596 & 0.746 \\
        & & Ours & \textbf{25.59} & 0.619 & \textbf{0.540} & \textbf{24.20} & \textbf{0.612} & \textbf{0.533} & \textbf{24.76} & \textbf{0.645} & \textbf{0.490} \\
        \bottomrule
    \end{tabular}
    \caption{Experimental results for Denoising, Image Fitting, and Super Resolution for \textbf{Baseline} model with \textbf{Ours}, evaluating their performance across three tasks. The best values in each column are highlighted in bold.}
    \label{supp-tab:results}
\end{table*}

\begin{table*}[t]
    \centering
    \begin{tabular}{l|cc|cc|cc}
    \toprule
        & \multicolumn{2}{c|}{\textbf{I-SIREN}} & \multicolumn{2}{c|}{\textbf{I-WIRE}} & \multicolumn{2}{c}{\textbf{I-Gauss}} \\
        & PSNR~$\uparrow$ & SSIM~$\uparrow$ & PSNR~$\uparrow$ & SSIM~$\uparrow$ & PSNR~$\uparrow$ & SSIM~$\uparrow$ \\
    \midrule
        Noise & \textbf{37.53} & \textbf{0.961} & \textbf{33.73} & \textbf{0.924} & \textbf{31.93} & \textbf{0.884} \\
        Ones  & 37.42 & 0.961 & 33.33 & 0.921 & 30.96 & 0.867 \\
        Zeros & 37.47 & 0.961 & 33.39 & 0.921 & 31.22 & 0.882 \\
    \bottomrule
    \end{tabular}
    \caption{Impact of different initial states (\( \mathcal{Z} \)) on image fitting task using the I-INR framework on Kodak dataset.}
    \label{tab:initialize}
\end{table*}

 \begin{table*}[t]
    \centering
    \begin{tabular}{c|cc|cc} 
        \toprule
        \textbf{Layers} & \multicolumn{2}{c|}{\textbf{I-SIREN}} & \multicolumn{2}{c}{\textbf{SIREN}} \\
        & PSNR & Params (k) & PSNR & Params (k) \\ 
        \midrule
        3 & 37.53 & 274 & 34.62 & 273 \\
        4 & 38.71 & 364 & 36.60 & 363 \\
        5 & 39.01 & 455 & 36.49 & 453 \\
        \bottomrule
    \end{tabular}
    \caption{Impact of varying Backbone network layer depths on the image fitting task using SIREN and I-SIREN on the Kodak dataset.} 
    \label{table:layers}
\end{table*}

\begin{table*}[t]
    \centering

    \begin{tabular}{l|ccc|ccc|ccc}
        \toprule
        & \multicolumn{3}{c|}{\textbf{I-SIREN}} & \multicolumn{3}{c|}{\textbf{I-WIRE}} & \multicolumn{3}{c}{\textbf{I-Gauss}} \\
        Fusion Method & PSNR$\uparrow$ & SSIM$\uparrow$ & LPIPS$\downarrow$ & PSNR$\uparrow$ & SSIM$\uparrow$ & LPIPS$\downarrow$ & PSNR$\uparrow$ & SSIM$\uparrow$ & LPIPS$\downarrow$ \\
        \midrule
        Multiplicative & \textbf{37.53} & \textbf{0.961} & \textbf{0.032} & \textbf{33.73} & \textbf{0.924} & \textbf{0.077} & \textbf{31.93} & \textbf{0.884} & \textbf{0.124} \\
        Adaptive & 37.01 & 0.960 & 0.037 & 33.63 & 0.923 & 0.078 & 31.80 & 0.883 & 0.129 \\
        \bottomrule
    \end{tabular}
    \caption{Comparison of fusion methods for the image-fitting task at $steps=2$ on the Kodak dataset with Gaussian initial state ($\mathcal{Z}\sim\mathcal{N}(0,1)$).}
    \label{tab:fusion_comparison_step2}
\end{table*}

\begin{table*}[t]
    \centering
    \begin{tabular}{c|c|c|c}
        \toprule
        \textbf{Method} & \textbf{Train Time (s)} & \textbf{Inference (ms / image)} & \textbf{PSNR (dB)} \\
        \midrule
        SIREN & 100 & 18.9 & 34.57 \\
        I-SIREN (2 steps) & 106 & 21.9 & \textbf{37.53} \\
        \bottomrule
    \end{tabular}
    \caption{Training and inference time comparison on the Kodak dataset (RTX\,4090).}
    \label{tab:time_comparison}
\end{table*}

\begin{table*}[t] 
    \centering

    \begin{tabular}{l|ccc} 
        \toprule
        \textbf{Method} & \textbf{Image Fitting} & \textbf{SR} & \textbf{Denoising} \\
        \midrule
        I-SIREN & 37.50 $\pm$ 0.247 & 27.22 $\pm$ 0.045 & 25.36 $\pm$ 0.044 \\
        SIREN   & 34.59 $\pm$ 0.027 & 26.77 $\pm$ 0.035 & 23.89 $\pm$ 0.021 \\
        \bottomrule
    \end{tabular}   
    \caption{Evaluation on 5 different seeds for image fitting, super-resolution tasks, and denoising on SIREN and I-SIREN.}
    \label{tab:std}
\end{table*}

\section{Training Setup}
Table~\ref{tab:multiplicative_combination} presents the key hyperparameters for various experiments. These hyperparameters were chosen to maximize peak performance across different tasks. 

Additionally, we used 2000 training iterations for image fitting, super-resolution (SR), and denoising tasks, while for object occupancy, we used 200 training iterations.

\begin{table*}[t]
    \centering
    \begin{tabular}{c|c|c|c|c}
        \toprule
        \textbf{Task} & \textbf{Parameter} & \textbf{I-Gauss} & \textbf{I-SIREN} & \textbf{I-WIRE} \\
        \midrule
        \multirow{2}{*}{Image Fitting} & $\omega$ & - & 52 & 7 \\
                                       & $\sigma$ & 18 & - & 13 \\
        \midrule
        \multirow{2}{*}{Image SR} & $\omega$ & - & 30 & 4 \\
                                  & $\sigma$ & 11 & - & 10 \\
        \midrule
        \multirow{2}{*}{Image Denoising} & $\omega$ & - & 55 & 10 \\
                                         & $\sigma$ & 18 & - & 16 \\
        \midrule
        \multirow{2}{*}{3D Occupancy} & $\omega$ & - & 55 & 10 \\
                                      & $\sigma$ & 17 & - & 20 \\
        \bottomrule
    \end{tabular}
    \caption{Hyperparameter settings for I-INRs across different tasks.}

    \label{tab:multiplicative_combination}
\end{table*}




\end{document}